# INTENTION-BASED SEGMENTATION:
# HUMAN RELIABILITY AND CORRELATION WITH LINGUISTIC CUES

**Rebecca J. Passonneau**
Department of Computer Science
Columbia University
New York, NY 10027
becky@cs.columbia.edu

**Diane J. Litman**
AT&T Bell Laboratories
600 Mountain Avenue
Murray Hill, NJ 07974
diane@research.att.com

## Abstract

Certain spans of utterances in a discourse, referred to here as *segments*, are widely assumed to form coherent units. Further, the segmental structure of discourse has been claimed to constrain and be constrained by many phenomena. However, there is weak consensus on the nature of segments and the criteria for recognizing or generating them. We present quantitative results of a two part study using a corpus of spontaneous, narrative monologues. The first part evaluates the statistical reliability of human segmentation of our corpus, where speaker intention is the segmentation criterion. We then use the subjects' segmentations to evaluate the correlation of discourse segmentation with three linguistic cues (referential noun phrases, cue words, and pauses), using information retrieval metrics.

## INTRODUCTION

A discourse consists not simply of a linear sequence of utterances,[1] but of meaningful relations among the utterances. As in much of the literature on discourse processing, we assume that certain spans of utterances, referred to here as *discourse segments*, form coherent units. The segmental structure of discourse has been claimed to constrain and be constrained by disparate phenomena: cue phrases (Hirschberg and Litman, 1993; Grosz and Sidner, 1986; Reichman, 1985; Cohen, 1984); lexical cohesion (Morris and Hirst, 1991); plans and intentions (Carberry, 1990; Litman and Allen, 1990; Grosz and Sidner, 1986); prosody (Grosz and Hirschberg, 1992; Hirschberg and Grosz, 1992; Hirschberg and Pierrehumbert, 1986); reference (Webber, 1991; Grosz and Sidner, 1986; Linde, 1979); and tense (Webber, 1988; Hwang and Schubert, 1992; Song and Cohen, 1991). However, there is weak consensus on the nature of segments and the criteria for recognizing or generating them in a natural language processing system. Until recently, little empirical work has been directed at establishing objectively verifiable segment boundaries, even though this is a precondition for

```
SEGMENT 1
Okay.
tsk There's  a farmer 
he looks like ay uh Chicano American,
he is picking pears.
A-nd u-m he's just picking them,
he comes off of the ladder,
a-nd he- u-h puts his pears into the basket.

    SEGMENT 2
    U-h a number of people are going by,
    and one is um /you know/ I don't know,
    I can't remember the first . . . the first person that goes by.
    Oh.
    A u-m a man with a goat comes by.
    It see it seems to be a busy place.
    You know,
    fairly busy,
    it's out in the country,
    maybe in u-m u-h the valley or something.

A-nd um  he  goes up the ladder,
and picks some more pears.
. . .
```

Figure 1: Discourse Segment Structure

avoiding circularity in relating segments to linguistic phenomena. We present the results of a two part study on the reliability of human segmentation, and correlation with linguistic cues. We show that human subjects can reliably perform discourse segmentation using speaker intention as a criterion. We use the segmentations produced by our subjects to quantify and evaluate the correlation of discourse segmentation with three linguistic cues: referential noun phrases, cue words, and pauses.

Figure 1 illustrates how discourse structure interacts with reference resolution in an excerpt taken from our corpus. The utterances of this discourse are grouped into two hierarchically structured segments, with segment 2 embedded in segment 1. This segmental structure is crucial for determining that the boxed pronoun *he* corefers with the boxed noun phrase *a farmer*. Without the segmentation, the referent of the underlined noun phrase *a man with a goat* is a potential referent of the pronoun because it is the most recent noun phrase consistent with the number and gender restrictions of the pronoun. With the segmentation analysis, *a man with a goat* is ruled out on structural grounds; this noun phrase occurs in segment 2, while the pronoun occurs after resumption of segment 1. *A farmer* is thus the most recent noun phrase that is both consistent with, and in the relevant interpretation context of, the pronoun in question.

---

[1] We use the term utterance to mean a use of a sentence or other linguistic unit, whether in text or spoken language.

One problem in trying to model such discourse structure effects is that segmentation has been observed to be rather subjective (Mann et al., 1992; Johnson, 1985). Several researchers have begun to investigate the ability of humans to agree with one another on segmentation. Grosz and Hirschberg (Grosz and Hirschberg, 1992; Hirschberg and Grosz, 1992) asked subjects to structure three AP news stories (averaging 450 words in length) according to the model of Grosz and Sidner (1986). Subjects identified hierarchical structures of discourse segments, as well as local structural features, using text alone as well as text and professionally recorded speech. Agreement ranged from 74%-95%, depending upon discourse feature. Hearst (1993) asked subjects to place boundaries between paragraphs of three expository texts (length 77 to 160 sentences), to indicate topic changes. She found agreement greater than 80%. We present results of an empirical study of a large corpus of spontaneous oral narratives, with a large number of potential boundaries per narrative. Subjects were asked to segment transcripts using an informal notion of speaker intention. As we will see, we found agreement ranging from 82%-92%, with very high levels of statistical significance (from $p = .114 \times 10^{-6}$ to $p \leq .6 \times 10^{-9}$).

One of the goals of such empirical work is to use the results to correlate linguistic cues with discourse structure. By asking subjects to segment discourse using a non-linguistic criterion, the correlation of linguistic devices with independently derived segments can be investigated. Grosz and Hirschberg (Grosz and Hirschberg, 1992; Hirschberg and Grosz, 1992) derived a discourse structure for each text in their study, by incorporating the structural features agreed upon by all of their subjects. They then used statistical measures to characterize these discourse structures in terms of acoustic-prosodic features. Morris and Hirst (1991) structured a set of magazine texts using the theory of Grosz and Sidner (1986). They developed a lexical cohesion algorithm that used the information in a thesaurus to segment text, then qualitatively compared their segmentations with the results. Hearst (1993) derived a discourse structure for each text in her study, by incorporating the boundaries agreed upon by the majority of her subjects. Hearst developed a lexical algorithm based on information retrieval measurements to segment text, then qualitatively compared the results with the structures derived from her subjects, as well as with those produced by Morris and Hirst. Iwanska (1993) compares her segmentations of factual reports with segmentations produced using syntactic, semantic, and pragmatic information. We derive segmentations from our empirical data based on the statistical significance of the agreement among subjects, or *boundary strength*. We develop three segmentation algorithms, based on results in the discourse literature. We use measures from information retrieval to quantify and evaluate the correlation between the segmentations produced by our algorithms and those derived from our subjects.

## RELIABILITY

The correspondence between discourse segments and more abstract units of meaning is poorly understood (see (Moore and Pollack, 1992)). A number of alternative proposals have been presented which directly or indirectly relate segments to intentions (Grosz and Sidner, 1986), RST relations (Mann et al., 1992) or other semantic relations (Polanyi, 1988). We present initial results of an investigation of whether naive subjects can reliably segment discourse using speaker intention as a criterion.

Our corpus consists of 20 narrative monologues about the same movie, taken from Chafe (1980) (N≈14,000 words). The subjects were introductory psychology students at the University of Connecticut and volunteers solicited from electronic bulletin boards. Each narrative was segmented by 7 subjects. Subjects were instructed to identify each point in a narrative where the speaker had completed one communicative task, and began a new one. They were also instructed to briefly identify the speaker's intention associated with each segment. Intention was explained in common sense terms and by example (details in (Litman and Passonneau, 1993)). To simplify data collection, we did not ask subjects to identify the type of hierarchical relations among segments illustrated in Figure 1. In a pilot study we conducted, subjects found it difficult and time-consuming to identify non-sequential relations. Given that the average length of our narratives is 700 words, this is consistent with previous findings (Rotondo, 1984) that non-linear segmentation is impractical for naive subjects in discourses longer than 200 words.

Since prosodic phrases were already marked in the transcripts, we restricted subjects to placing boundaries between prosodic phrases. In principle, this makes it more likely that subjects will agree on a given boundary than if subjects were completely unrestricted. However, previous studies have shown that the smallest unit subjects use in similar tasks corresponds roughly to a breath group, prosodic phrase, or clause (Chafe, 1980; Rotondo, 1984; Hirschberg and Grosz, 1992). Using smaller units would have artificially lowered the probability for agreement on boundaries.

Figure 2 shows the responses of subjects at each potential boundary site for a portion of the excerpt from Figure 1. Prosodic phrases are numbered sequentially, with the first field indicating prosodic phrases with sentence-final contours, and the second

```
3.3  [.35+ [.35] a-nd] he- u-h [.3] puts his pears into the basket.
     ┌──────────┐
     │6 SUBJECTS│                                        NP, PAUSE
     └──────────┘
4.1  [1.0 [.5] U-h] a number of people are going by,
                                                       CUE, PAUSE
4.2  [.35+ and [.35]] one is [1.15 um] /you know/ I don't know,

4.3  I can't remember the first ... the first person that goes by.
     ┌──────────┐
     │1 SUBJECTS│                                             PAUSE
     └──────────┘
5.1  [.3] Oh.
     ┌──────────┐
     │1 SUBJECTS│                                                NP
     └──────────┘
6.1  A u-m.. a man with a goat [.2] comes by.
     ┌──────────┐
     │2 SUBJECTS│                                        NP, PAUSE
     └──────────┘
7.1  [.25] It see it seems to be a busy place.
                                                              PAUSE
8.1  [.1] You know,

8.2  fairly busy,
     ┌──────────┐
     │1 SUBJECTS│
     └──────────┘
8.3  it's out in the country,
                                                              PAUSE
8.4  [.4] maybe in u-m [.8] u-h the valley or something.
     ┌──────────┐
     │7 SUBJECTS│                                   NP, CUE, PAUSE
     └──────────┘
9.1  [2.95 [.9] A-nd um [.25] [.35]] he goes up the ladder,
```

Figure 2: Excerpt from 9, with Boundaries

field indicating phrase-final contours.[2] Line spaces *between* prosodic phrases represent potential boundary sites. Note that a majority of subjects agreed on only 2 of the 11 possible boundary sites: after 3.3 (n=6) and after 8.4 (n=7). (The symbols NP, CUE and PAUSE will be explained later.)

Figure 2 typifies our results. Agreement among subjects was far from perfect, as shown by the presence here of 4 boundary sites identified by only 1 or 2 subjects. Nevertheless, as we show in the following sections, the degree of agreement among subjects is high enough to demonstrate that segments can be reliably identified. In the next section we discuss the percent agreement among subjects. In the subsequent section we show that the frequency of boundary sites where a majority of subjects assign a boundary is highly significant.

## AGREEMENT AMONG SUBJECTS

We measure the ability of subjects to agree with one another, using a figure called percent agreement. *Percent agreement*, defined in (Gale et al., 1992), is the ratio of observed agreements with the majority opinion to possible agreements with the majority opinion. Here, agreement among four, five, six, or seven subjects on whether or not there is a segment boundary between two adjacent prosodic phrases constitutes a *majority opinion*. Given a transcript of length n prosodic phrases, there are n-1 possible boundaries. The total *possible agreements* with the majority corresponds to the number of subjects times n-1. Total *observed agreements* equals the number of times that subjects' boundary decisions agree with the majority opinion. As noted above, only 2 of the 11 possible boundaries in Figure 2 are boundaries using the majority opinion criterion. There are 77 possible agreements with the majority opinion, and 71 observed agreements. Thus, percent agreement for the excerpt as a whole is 71/77, or 92%. The breakdown of agreement on boundary and non-boundary majority opinions is 13/14 (93%) and 58/63 (92%), respectively.

The figures for percent agreement with the majority opinion for all 20 narratives are shown in Table 1. The columns represent the narratives in our corpus. The first two rows give the absolute number of potential boundary sites in each narrative (i.e., n-1) followed by the corresponding percent agreement figure for the narrative as a whole. Percent agreement in this case averages 89% (variance $\sigma$=.0006; max.=92%; min.=82%). The next two pairs of rows give the figures when the majority opinions are broken down into boundary and non-boundary opinions, respectively. Non-boundaries, with an average percent agreement of 91% ($\sigma$=.0006; max.=95%; min.=84%), show greater agreement among subjects than boundaries, where average percent agreement is 73% ($\sigma$= .003; max.=80%; min.=60%). This partly reflects the fact that non-boundaries greatly outnumber boundaries, an average of 89 versus 11 majority opinions per transcript. The low variances, or spread around the average, show that subjects are also consistent with one another.

Defining a task so as to maximize percent agreement can be difficult. The high and consistent levels of agreement for our task suggest that we have found a useful experimental formulation of the task of discourse segmentation. Furthermore, our percent agreement figures are comparable with the results of other segmentation studies discussed above. While studies of other tasks have achieved stronger results (e.g., 96.8% in a word-sense disambiguation study (Gale et al., 1992)), the meaning of percent agreement in isolation is unclear. For example, a percent agreement figure of less than 90% could still be very meaningful if the probability of obtaining such a figure is low. In the next section we demonstrate the significance of our findings.

## STATISTICAL SIGNIFICANCE

We represent the segmentation data for each narrative as an $i \times j$ matrix of height i=7 subjects and width j=n-1. The value in each cell $c_{i,j}$ is a one if the *i*th subject assigned a boundary at site *j*, and a zero if they did not. We use Cochran's test (Cochran, 1950) to evaluate significance of differences across columns in the matrix.[3]

Cochran's test assumes that the number of 1s within a single row of the matrix is fixed by observation, and that the totals across rows can vary. Here a row total corresponds to the total number of boundaries assigned by subject i. In the case of narrative 9 (j=96), one of the subjects assigned 8 boundaries. The probability of a 1 in any of the j cells of the row is thus 8/96, with $\binom{96}{8}$ ways for the 8 boundaries to be distributed. Taking this into account for each row, Cochran's test evaluates the null hypothesis that the number of 1s in a column, here the total number of subjects assigning a boundary at the jth site, is randomly distributed. Where the row totals are $u_i$, the column totals are $T_j$, and the average column total is $\overline{T}$, the statistic is given by:

$$Q = \frac{j\,(j-1)\sum (T_j - \overline{T})^2}{c\,(\sum u_i) - (\sum u_i^2)}$$

---

[2]The transcripts presented to subjects did not contain line numbering or pause information (pauses indicated here by bracketed numbers.)

[3]We thank Julia Hirschberg for suggesting this test.

| Narrative    | 1   | 2   | 3  | 4  | 5  | 6  | 7  | 8  | 9  | 10 | 11  | 12  | 13  | 14  | 15  | 16  | 17 | 18  | 19 | 20 |
|---|---|---|---|---|---|---|---|---|---|---|---|---|---|---|---|---|---|---|---|---|
| All Opinions | 138 | 121 | 55 | 63 | 69 | 83 | 90 | 50 | 96 | 195 | 110 | 160 | 108 | 113 | 112 | 46 | 151 | 85 | 94 | 56 |
| % Agreement  | 87  | 82  | 91 | 89 | 89 | 90 | 90 | 90 | 90 | 88  | 92  | 90  | 91  | 89  | 85  | 89 | 92  | 91 | 91 | 86 |
| Boundary     | 21  | 16  | 7  | 10 | 6  | 5  | 11 | 5  | 8  | 22  | 13  | 17  | 9   | 11  | 8   | 7  | 15  | 11 | 10 | 6  |
| % Agreement  | 74  | 70  | 76 | 77 | 60 | 80 | 79 | 69 | 75 | 70  | 74  | 75  | 73  | 71  | 68  | 73 | 77  | 71 | 80 | 74 |
| Non-Boundary | 117 | 105 | 48 | 53 | 63 | 78 | 79 | 45 | 88 | 173 | 97  | 143 | 99  | 102 | 104 | 39 | 136 | 74 | 84 | 50 |
| % Agreement  | 89  | 84  | 93 | 91 | 92 | 91 | 92 | 92 | 92 | 90  | 95  | 91  | 93  | 91  | 87  | 92 | 93  | 94 | 93 | 88 |

Table 1: Percent Agreement with the Majority Opinion

$Q$ approximates the $\chi^2$ distribution with j-1 degrees of freedom (Cochran, 1950). Our results indicate that the agreement among subjects is extremely highly significant. That is, the number of 0s or 1s in certain columns is much greater than would be expected by chance. For the 20 narratives, the probabilities of the observed distributions range from $p = .114 \times 10^{-6}$ to $p \leq .6 \times 10^{-9}$.

The percent agreement analysis classified all the potential boundary sites into two classes, boundaries versus non-boundaries, depending on how the majority of subjects responded. This is justified by further analysis of $Q$. As noted in the preceding section, the proportion of *non*-boundaries agreed upon by most subjects (i.e., where $0 \leq T_j \leq 3$) is higher than the proportion of *boundaries* they agree on ($4 \leq T_j \leq 7$). That agreement on non-boundaries is more probable suggests that the significance of $Q$ owes most to the cases where columns have a majority of 1's. This assumption is borne out when $Q$ is partitioned into distinct components for each possible value of $T_j$ (0 to 7), based on partioning the sum of squares in the numerator of $Q$ into distinct samples (Cochran, 1950). We find that $Q_j$ is significant for each distinct $T_j \geq 4$ across all narratives. For $T_j=4$, $.0002 \leq p \leq .30 \times 10^{-8}$; probabilities become more significant for higher levels of $T_j$, and the converse. At $T_j=3$, p is sometimes above our significance level of .01, depending on the narrative.

## DISCUSSION OF RESULTS

We have shown that an atheoretical notion of speaker intention is understood sufficiently uniformly by naive subjects to yield significant agreement across subjects on segment boundaries in a corpus of oral narratives. We obtained high levels of percent agreement on boundaries as well as on non-boundaries. Because the average narrative length is 100 prosodic phrases and boundaries are relatively infrequent (average boundary frequency=16%), percent agreement among 7 subjects (row one in Table 1) is largely determined by percent agreement on non-boundaries (row three). Thus, total percent agreement could be very high, even if subjects did not agree on any boundaries. However, our results show that percent agreement on boundaries is not only high (row two), but also statistically significant.

We have shown that boundaries agreed on by at least 4 subjects are very unlikely to be the result of chance. Rather, they most likely reflect the validity of the notion of segment as defined here. In Figure 2, 6 of the 11 possible boundary sites were identified by at least 1 subject. Of these, only two were identified by a majority of subjects. If we take these two boundaries, appearing after prosodic phrases 3.3 and 8.4, to be statistically validated, we arrive at a linear version of the segmentation used in Figure 1. In the next section we evaluate how well statistically validated boundaries correlate with the distribution of linguistic cues.

## CORRELATION

In this section we present and evaluate three discourse segmentation algorithms, each based on the use of a single linguistic cue: referential noun phrases (NPs), cue words, and pauses.[4] While the discourse effects of these and other linguistic phenomena have been discussed in the literature, there has been little work on examining the use of such effects for recognizing or generating segment boundaries,[5] or on evaluating the comparative utility of different phenomena for these tasks. The algorithms reported here were developed based on ideas in the literature, then evaluated on a representative set of 10 narratives. Our results allow us to directly compare the performance of the three algorithms, to understand the utility of the individual knowledge sources.

We have not yet attempted to create comprehensive algorithms that would incorporate all possible relevant features. In subsequent phases of our work, we will tune the algorithms by adding and refining features, using the initial 10 narratives as a training set. Final evaluation will be on a test set corresponding to the 10 remaining narratives. The initial results reported here will provide us with a baseline for quantifying improvements resulting from distinct modifications to the algorithms.

---

[4] The input to each algorithm is a discourse transcription labeled with prosodic phrases. In addition, for the NP algorithm, noun phrases need to be labeled with anaphoric relations. The pause algorithm requires pauses to be noted.

[5] A notable exception is the literature on pauses.

|            | Subjects |              |
|------------|----------|--------------|
| Algorithm  | Boundary | Non-Boundary |
| Boundary   | a        | b            |
| Non-Boundary | c      | d            |

| Recall  | Precision | Fallout | Error         |
|---------|-----------|---------|---------------|
| a/(a+c) | a/(a+b)   | b/(b+d) | (b+c)/(a+b+c+d) |

Table 2: Evaluation Metrics

We use metrics from the area of information retrieval to evaluate the performance of our algorithms. The correlation between the boundaries produced by an algorithm and those independently derived from our subjects can be represented as a matrix, as shown in Table 2. The value $a$ (in cell $c_{1,1}$) represents the number of potential boundaries identified by both the algorithm and the subjects, $b$ the number identified by the algorithm but not the subjects, $c$ the number identified by the subjects but not the algorithm, and $d$ the number neither the algorithm nor the subjects identified. Table 2 also shows the definition of the four evaluation metrics in terms of these values. *Recall* errors represent the false rejection of a boundary, while *precision* errors represent the false acceptance of a boundary. An algorithm with perfect performance segments a discourse by placing a boundary at all and only those locations with a subject boundary. Such an algorithm has 100% recall and precision, and 0% fallout and error.

For each narrative, our human segmentation data provides us with a set of boundaries classified by 7 levels of subject strength: ($1 \leq T_j \leq 7$). That is, boundaries of strength 7 are the set of possible boundaries identified by all 7 subjects. As a baseline for examining the performance of our algorithms, we compare the boundaries produced by the algorithms to boundaries of strength $T_j \geq 4$. These are the statistically validated boundaries discussed above, i.e., those boundaries identified by 4 or more subjects. Note that recall for $T_j \geq 4$ corresponds to percent agreement for boundaries. We also examine the evaluation metrics for each algorithm, cross-classified by the individual levels of boundary strength.

**REFERENTIAL NOUN PHRASES**

Our procedure for encoding the input to the referring expression algorithm takes 4 factors into account, as documented in (Passonneau, 1993a). Briefly, we construct a 4-tuple for each referential NP: <FIC, NP, i, I>. FIC is clause location, NP is surface form, i is referential identity, and I is a set of inferential relations. Clause location is determined by sequentially assigning distinct indices to each functionally independent clause (FIC); an FIC is roughly equivalent to a tensed clause that is neither a verb argument nor a restrictive relative.

```
25    16.1   You could hear the bicycle_{12},
      16.2   wheels_{13} going round.
CODING       <25, wheels, 13, (13 r1 12)>
```

Figure 3: Sample Coding (from Narrative 4)

Figure 3 illustrates the coding of an NP, *wheels*. It's location is FIC number 25. The surface form is the string *wheels*. The wheels are new to the discourse, so the referential index 13 is new. The inferential relation (13 r1 12) indicates that the wheels entity is related to the bicycle entity (index 12) by a part/whole relation.[6]

The input to the segmentation algorithm is a list of 4-tuples representing all the referential NPs in a narrative. The output is a set of boundaries B, represented as ordered pairs of adjacent clauses: ($FIC_n$, $FIC_{n+1}$). Before describing how boundaries are assigned, we explain that the potential boundary locations for the algorithm, between each FIC, differ from the potential boundary locations for the human study, between each prosodic phrase. Cases where multiple prosodic phrases map to one FIC, as in Figure 3, simply reflect the use of additional linguistic features to reject certain boundary sites, e.g., (16.1,16.2). However, the algorithm has the potential to assign multiple boundaries between adjacent prosodic phrases. The example shown in Figure 4 has one boundary site available to the human subjects, between 3.1 and 3.2. Because 3.1 consists of multiple FICs (6 and 7) the algorithm can and does assign 2 boundaries here: (6,7) and (7,8). To normalize the algorithm output, we reduce multiple boundaries at a boundary site to one, here (7,8). A total of 5 boundaries are eliminated in 3 of the 10 test narratives (out of 213 in all 10). All the remaining boundaries (here (3.1,3.2)) fall into class $b$ of Table 2.

The algorithm operates on the principle that if an NP in the current FIC provides a referential link to the current segment, the current segment continues. However, NPs and pronouns are treated differently based on the notion of focus (cf. (Passonneau, 1993a). A third person definite pronoun provides a referential link if its index occurs anywhere in the *current segment*. Any other NP type provides a referential link if its index occurs in the *immediately preceding FIC*.

The symbol NP in Figure 2 indicates boundaries assigned by the algorithm. Boundary (3.3,4.1) is assigned because the sole NP in 4.1, *a number of people*, refers to a new entity, one that cannot be inferred from any entity mentioned in 3.3. Boundary

---

[6]We use 5 inferrability relations (Passonneau, 1993a). Since there is a phrase boundary between *the bicycle* and *wheels*, we do not take *bicycle* to modify *wheels*.

```
6   3.1   A-nd he's not ... paying all that much attention
              NP BOUNDARY
7         because you know the pears fall,
              NP BOUNDARY      (no subjects)
8   3.2   and he doesn't really notice,
```
Figure 4: Multiple FICs in One Prosodic Phrase

FORALL $FIC_n, 1 \leq n \leq last$

IF $CD_n \cap CD_{n-1} \neq \emptyset$ THEN $CD_S = CD_S \cup CD_n$
% (COREFERENTIAL LINK TO NP IN $FIC_{n-1}$)

ELSE IF $F_n \cap CD_{n-1} \neq \emptyset$ THEN $CD_S = CD_S \cup CD_n$
% (INFERENTIAL LINK TO NP IN $FIC_{n-1}$)

ELSE IF $PRO_n \cap CD_S \neq \emptyset$ THEN $CD_S = CD_S \cup CD_n$
% (DEFINITE PRONOUN LINK TO SEGMENT)

ELSE $B = B \cup \{(FIC_{n-1}, FIC_n)\}$
% (IF NO LINK, ADD A BOUNDARY)

Figure 5: Referential NP Algorithm

(8.4,9.1) results from the following facts about the NPs in 9.1: 1) the full NP *the ladder* is not referred to implicitly or explicitly in 8.4, 2) the third person pronoun *he* refers to an entity, the farmer, that was last mentioned in 3.3, and 3 NP boundaries have been assigned since then. If the farmer had been referred to anywhere in 7.1 through 8.4, no boundary would be assigned at (8.4,9.1).

Figure 5 illustrates the three decision points of the algorithm. $FIC_n$ is the current clause (at location n); $CD_n$ is the set of all indices for NPs in $FIC_n$; $F_n$ is the set of entities that are inferentially linked to entities in $CD_n$; $PRO_n$ is the subset of $CD_n$ where NP is a third person definite pronoun; $CD_{n-1}$ is the contextual domain for the previous FIC, and $CD_S$ is the contextual domain for the current segment. $FIC_n$ continues the current segment if it is anaphorically linked to the preceding clause 1) by a coreferential NP, or 2) by an inferential relation, or 3) if a third person definite pronoun in $FIC_n$ refers to an entity in the current segment. If no boundary is added, $CD_S$ is updated with $CD_n$. If all 3 tests fail, $FIC_n$ is determined to begin a new segment, and $(FIC_{n-1}, FIC_n)$ is added to B.

Table 3 shows the average performance of the referring expression algorithm (row labelled NP) on the 4 measures we use here. Recall is .66 ($\sigma$=.068; max=1; min=.25), precision is .25 ($\sigma$=.013; max=.44; min=.09), fallout is .16 ($\sigma$=.004) and error rate is 0.17 ($\sigma$=.005). Note that the error rate and fallout, which in a sense are more sensitive measures of inaccuracy, are both much lower than the precision and have very low variance. Both recall and precision have a relatively high variance.

## CUE WORDS

Cue words (e.g., "now") are words that are sometimes used to explicitly signal the structure of a discourse. We develop a baseline segmentation algorithm based on cue words, using a simplification of one of the features shown by Hirschberg and Litman (1993) to identify discourse usages of cue words. Hirschberg and Litman (1993) examine a large set of cue words proposed in the literature and show that certain prosodic and structural features, including a position of first in prosodic phrase, are highly correlated with the discourse uses of these words. The input to our lower bound cue word algorithm is a sequential list of the prosodic phrases constituting a given narrative, the same input our subjects received. The output is a set of boundaries B, represented as ordered pairs of adjacent phrases ($P_n, P_{n+1}$), such that the first item in $P_{n+1}$ is a member of the set of cue words summarized in Hirschberg and Litman (1993). That is, if a cue word occurs at the beginning of a prosodic phrase, the usage is assumed to be discourse and thus the phrase is taken to be the beginning of a new segment. Figure 2 shows 2 boundaries (CUE) assigned by the algorithm, both due to *and*.

Table 3 shows the average performance of the cue word algorithm for statistically validated boundaries. Recall is 72% ($\sigma$=.027; max=.88; min=.40), precision is 15% ($\sigma$=.003; max=.23; min=.04), fallout is 53% ($\sigma$=.006) and error is 50% ($\sigma$=.005). While recall is quite comparable to human performance (row 4 of the table), the precision is low while fallout and error are quite high. Precision, fallout and error have much lower variance, however.

## PAUSES

Grosz and Hirschberg (Grosz and Hirschberg, 1992; Hirschberg and Grosz, 1992) found that in a corpus of recordings of AP news texts, phrases beginning discourse segments are correlated with duration of preceding pauses, while phrases ending discourse segments are correlated with subsequent pauses. We use a simplification of these results to develop a baseline algorithm for identifying boundaries in our corpus using pauses. The input to our pause segmentation algorithm is a sequential list of all prosodic phrases constituting a given narrative, with pauses (and their durations) noted. The output is a set of boundaries B, represented as ordered pairs of adjacent phrases ($P_n, P_{n+1}$), such that there is a pause between $P_n$ and $P_{n+1}$. Unlike Grosz and Hirschberg, we do not currently take phrase duration into account. In addition, since our segmentation task is not hierarchical, we do not note whether phrases begin, end, suspend, or resume segments. Figure 2 shows boundaries (PAUSE) assigned by the algorithm.

Table 3 shows the average performance of the pause algorithm for statistically validated boundaries. Recall is 92% ($\sigma$=.008; max=1; min=.73), precision is 18% ($\sigma$=.002; max=.25; min=.09), fallout is 54% ($\sigma$=.004), and error is 49% ($\sigma$=.004). Our algorithm thus performs with recall higher than human performance. However, precision is low,

|        | Recall | Precision | Fallout | Error |
|--------|--------|-----------|---------|-------|
| NP     | .66    | .25       | .16     | .17   |
| Cue    | .72    | .15       | .53     | .50   |
| Pause  | .92    | .18       | .54     | .49   |
| Humans | .74    | .55       | .09     | .11   |

Table 3: Evaluation for $T_j \geq 4$

| $T_j$     | 1    | 2    | 3    | 4   | 5   | 6   | 7   |
|-----------|------|------|------|-----|-----|-----|-----|
| N         | 40.3 | 17.9 | 11.9 | 8.2 | 5.7 | 4.5 | 3.4 |
| NPs       |      |      |      |     |     |     |     |
| Recall    | .20  | .31  | .37  | .12 | .62 | .52 | .48 |
| Precision | .18  | .26  | .15  | .02 | .15 | .07 | .06 |
| Cues      |      |      |      |     |     |     |     |
| Recall    | .54  | .60  | .76  | .82 | .75 | .59 | .65 |
| Precision | .17  | .09  | .08  | .07 | .04 | .03 | .02 |
| Pauses    |      |      |      |     |     |     |     |
| Recall    | .55  | .65  | .73  | .84 | .95 | .97 | .98 |
| Precision | .18  | .10  | .08  | .06 | .06 | .04 | .03 |
| Humans    |      |      |      |     |     |     |     |
| Recall    | .14  | .29  | .43  | .57 | .71 | .86 | 1.0 |
| Precision | .14  | .14  | .17  | .15 | .15 | .13 | .14 |

Table 4: Variation with Boundary Strength

while both fallout and error are quite high.

## DISCUSSION OF RESULTS

In order to evaluate the performance measures for the algorithms, it is important to understand how individual humans perform on all 4 measures. Row 4 of Table 3 reports the average individual performance for the 70 subjects on the 10 narratives. The average recall for humans is .74 ($\sigma$=.038),[7] and the average precision is .55 ($\sigma$=.027), much lower than the ideal scores of 1. The fallout and error rates of .09 ($\sigma$=.004) and .11 ($\sigma$=.003) more closely approximate the ideal scores of 0. The low recall and precision reflect the considerable variation in the number of boundaries subjects assign, as well as the imperfect percent agreement (Table 1).

To compare algorithms, we must take into account the dimensions along which they differ apart from the different cues. For example, the referring expression algorithm (RA) differs markedly from the pause and cue algorithms (PA, CA) in using more knowledge. CA and PA depend only on the ability to identify boundary sites, potential cue words and pause locations while RA relies on 4 features of NPs to make 3 different tests (Figure 5). Unsurprisingly, RA performs most like humans. For both CA and PA, the recall is relatively high, but the precision is very low, and the fallout and error rate are both very high. For RA, recall and precision are not as different, precision is higher than CA and PA, and fallout and error rate are both relatively low.

A second dimension to consider in comparing performance is that humans and RA assign boundaries based on a global criterion, in contrast to CA and PA. Subjects typically use a relatively gross level of speaker intention. By default, RA assumes that the current segment continues, and assigns a boundary under relatively narrow criteria. However, CA and PA rely on cues that are relevant at the local as well as the global level, and consequently assign boundaries more often. This leads to a preponderance of cases where PA and CA propose a boundary but where a majority of humans did not, category $b$ from Table 2. High $b$ lowers precision, reflected in the low precision for CA and PA.

We are optimistic that all three algorithms can be improved, for example, by discriminating among types of pauses, types of cue words, and features of referential NPs. We have enhanced RA with certain grammatical role features following (Passonneau, 1993b). In a preliminary experiment using boundaries from our first set of subjects (4 per narrative instead of 7), this increased both recall and precision by $\sim 10\%$.

The statistical results validate boundaries agreed on by a majority of subjects, but do not thereby invalidate boundaries proposed by only 1-3 subjects. We evaluate how performance varies with boundary strength ($1 \leq T_j \leq 7$). Table 4 shows recall and precision of RA, PA, CA and humans when boundaries are broken down into those identified by exactly 1 subject, exactly 2, and so on up to 7.[8] There is a strong tendency for recall to increase and precision to decrease as boundary strength increases. We take this as evidence that the presence of a boundary is not a binary decision; rather, that boundaries vary in perceptual salience.

## CONCLUSION

We have shown that human subjects can reliably perform linear discourse segmentation in a corpus of transcripts of spoken narratives, using an informal notion of speaker intention. We found that percent agreement with the segmentations produced by the majority of subjects ranged from 82%-92%, with an average across all narratives of 89% ($\sigma$=.0006). We found that these agreement results were highly significant, with probabilities of randomly achieving our findings ranging from $p = .114 \times 10^{-6}$ to $p \leq .6 \times 10^{-9}$.

We have investigated the correlation of our intention-based discourse segmentations with referential noun phrases, cue words, and pauses. We developed segmentation algorithms based on the use of each of these linguistic cues, and quantitatively evaluated their performance in identifying the statistically validated boundaries independently produced by our subjects. We found that compared to human performance, the recall of the three algorithms was comparable, the precision was much lower, and the fallout and error of only the noun phrase algorithm was comparable. We also found a tendency for recall to increase and precision to decrease with exact boundary strength, suggesting that the cognitive salience of boundaries is graded.

---

[7] Human recall is equivalent to percent agreement for boundaries. However, the average shown here represents only 10 narratives, while the average from Table 1 represents all 20.

[8] Fallout and error rate do not vary much across $T_j$.

While our initial results are promising, there is certainly room for improvement. In future work on our data, we will attempt to maximize the correlation of our segmentations with linguistic cues by improving the performance of our individual algorithms, and by investigating ways to combine our algorithms (cf. Grosz and Hirschberg (1992)). We will also explore the use of alternative evaluation metrics (e.g. string matching) to support close as well as exact correlation.

## ACKNOWLEDGMENTS

The authors wish to thank W. Chafe, K. Church, J. DuBois, B. Gale, V. Hatzivassiloglou, M. Hearst, J. Hirschberg, J. Klavans, D. Lewis, E. Levy, K. McKeown, E. Siegel, and anonymous reviewers for helpful comments, references and resources. Both authors' work was partially supported by DARPA and ONR under contract N00014-89-J-1782; Passonneau was also partly supported by NSF grant IRI-91-13064.